\documentclass[runningheads]{llncs}
\usepackage{graphicx}
\usepackage{amsmath,amssymb} % define this before the line numbering.
\usepackage{color}
\usepackage{caption} 
\usepackage{hyperref}

\newcommand{\etal}{\textit{et al.}}

\begin{document}
\pagestyle{headings}
\mainmatter

\def\papertitle{A Comprehensive Review of Skeleton-based Movement Assessment Methods}

\def\ACCV20SubNumber{***}  % Insert your submission number here

%===========================================================
\title{\papertitle} % Replace with your title
\titlerunning{\papertitle}
\authorrunning{Tal Hakim}

\author{Tal Hakim \\
{\tt\small thakim.research@gmail.com}}

\institute{}

\maketitle

%===========================================================
\begin{abstract}
The raising availability of 3D cameras and dramatic improvement of computer vision algorithms in the recent decade, accelerated the research of automatic movement assessment solutions. Such solutions can be implemented at home, using affordable equipment and dedicated software. In this paper, we divide the movement assessment task into secondary tasks and explain why they are needed and how they can be addressed. We review the recent solutions for automatic movement assessment from skeleton videos, comparing them by their objectives, features, movement domains and algorithmic approaches. In addition, we discuss the status of the research on this topic in a high level.

\end{abstract}

%===========================================================
\section{Introduction}
One of the most significant incentives for recent research on movement assessment, is the availability of affordable 3D skeleton recognition devices, such as Kinect, which redefine the target audience of applications that are based on user pose and movement. Addressing this problem is considered a hard task, as it requires paying attention to timings, performances and low-level details.

In the recent decade, different solutions have been proposed for dealing with automatic assessment of movements, based on machine-learning algorithms. In this work, we review these solutions and compare them.

We divide the assessment problem into two typical problems of detecting abnormalities in repetitive movements and predicting scores in structured movements. We then list the existing works and their features and the existing public datasets. We elaborate on the secondary problems that take part in the algorithmic flow of typical movement assessment solutions and list the methods and algorithms used by the different works. Finally, we discuss the findings in a high level.

The outline of this review is as follows. In the next chapter, we first present the main types of movement assessment problems, list the features of existing works and list the used public datasets. In addition, we elaborate on the secondary problems and list the methods that were implemented to solve them. The last two chapters include a discussion and conclusions, respectively.

\section{Movement Assessment}
\label{lbl:movementAssessment}
\begin{figure}[]
\centering
  \includegraphics[width=0.75\linewidth,keepaspectratio]{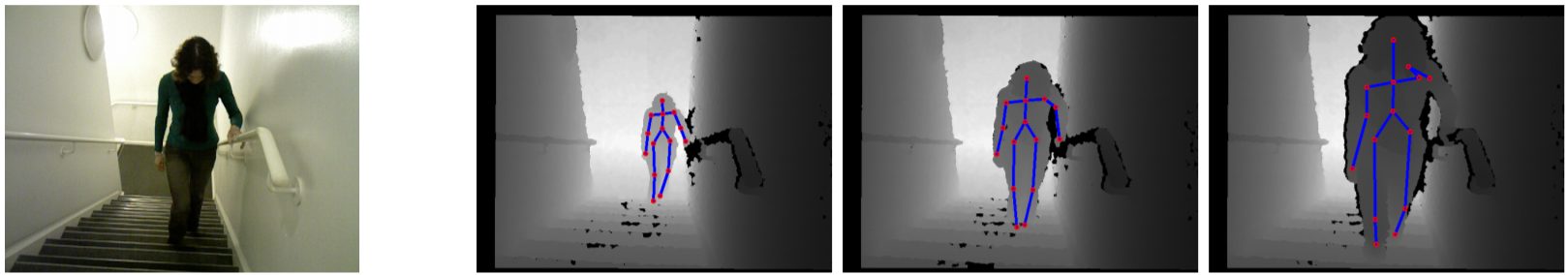}
  \caption[]{A walking-up-stairs movement with 3D skeleton joints detected from a Kinect RGB-D video~\cite{paiement2014online}.}
  \label{fig:gait}
\end{figure}

\begin{figure}[]
\centering
  \includegraphics[width=0.75\linewidth,keepaspectratio]{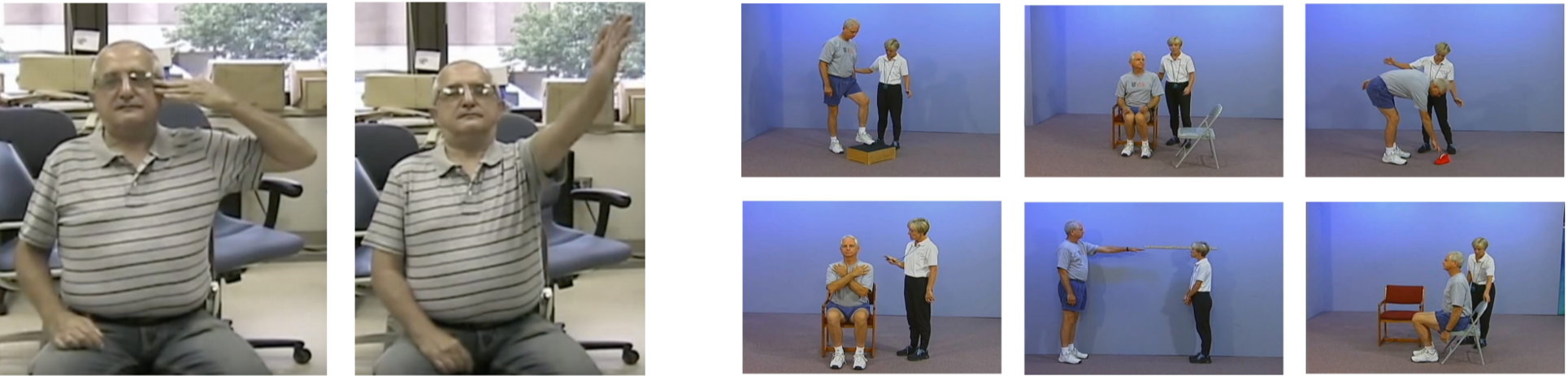}
  \caption[]{An FMA assessment~\cite{eichler2018non} and a BBS assessment~\cite{masalha2020predicting}.}
  \label{fig:fma_and_bbs}
\end{figure}

There are generally two main types of movement assessment solutions. The first type focuses on detecting abnormalities in relatively long, repetitive movements~\cite{paiement2014online,jun2020feature,chaaraoui2015abnormal,nguyen2016skeleton,devanne2016learning,baptista2018deformation,nguyen2018estimating,nguyen2018skeleton,khokhlova2018kinematic}, such as gait, as visualized in Figure~\ref{fig:gait}. The second type of movements, on the other hand, focuses on assessing structured movements~\cite{parisi2016human,su2013personal,eichler20183d,eichler2018non,hakim2019mal,hakim2019malf,masalha2020predicting,dressler2019data,dressler2020towards,hu2014real,capecci2016physical,capecci2018hidden,osgouei2018objective,al2019quantifying,cao2019novel,williams2019assessment,yu2019dynamic,lei2020learning}, such as movements from the Fugl-Meyer Assessment (FMA)~\cite{fugl1975post} or Berg Balance Scale (BBS)~\cite{bbs} medical assessments, as visualized in Figure~\ref{fig:fma_and_bbs}, which usually have clear definitions of starting positions, ending positions, objectives and constraints.

While most of the works deal with assessing known-in-advance, limited range of movement types, only a few works try to provide general solutions, which aim to be adaptive to new types of movements. Such works, which were evaluated on multiple types of movements~\cite{parisi2016human,su2013personal,eichler20183d,eichler2018non,hakim2019mal,hakim2019malf,masalha2020predicting,hu2014real,capecci2016physical,capecci2018hidden,al2019quantifying,cao2019novel,williams2019assessment,lei2020learning}, may therefore assume no prior knowledge on a learned movement type, such that they may need to automatically extract its most important properties from the training set, or use learning algorithms that are adaptive in their nature.

A typical movement assessment algorithm will need to address the following fundamental problems: capturing or detecting human skeleton joint positions, geometric normalization, temporal alignment, feature extraction, score prediction and feedback generation.  In this chapter, we review the solutions existing works implemented for each of these problems.

\subsection{Movement Domains and Solution Features}
Most of the works that deal with structured movements, mainly deal with predicting the quality of a performance and sometimes producing feedback. On the contrary, most of the works that deal with repetitive movements, such as gait, give more focus to detecting abnormalities and computing scores that are based on similarity to normal movements.

Table~\ref{tbl:features} summarizes the features of each of the works that deal with structured movements. When a solution produces a quality score on a continuous scale, then we consider the numerical score feature as existing. When a solution classifies performances into a discrete scale of qualities, then we consider the quality classification feature as existing.  When a solution produces unbound textual feedback or presents describable features that can be directly translated into textual feedback, then we consider the unbound feedback feature as existing. When a training set that only consists of proper performances is sufficient for a solution to work, then we consider the trains-on-proper-movements feature as existing.

\begin{table}

\centering
\resizebox{0.98\linewidth}{!}{%
\begin{tabular}{ |c|c|c|c|c|c|c| } 
 \hline
 \textbf{} & \textbf{Movement} & \textbf{No. Movement} & \textbf{Numerical} & \textbf{Quality} & \textbf{Unbound} & \textbf{Trains on} \\
 \textbf{Work} & \textbf{Domain} & \textbf{Types Evaluated} & \textbf{Score} &  \textbf{Classification} & \textbf{Feedback} & \textbf{Proper Movements} \\
 \hline
 \cite{parisi2016human} & Powerlifting  & 3 & \checkmark & & \checkmark & \checkmark \\
 \hline
 \cite{su2013personal} & Rehabilitation  & - & \checkmark & \checkmark & & \checkmark \\
 \hline
 \cite{eichler20183d,eichler2018non} & FMA  & 2 &  & \checkmark & & \\
 \hline
 \cite{hakim2019mal,hakim2019malf} & FMA & 3 & \checkmark & \checkmark & \checkmark & \checkmark \\
 \hline
 \cite{masalha2020predicting} & BBS  & 14 &   & \checkmark & & \\
 \hline
 \cite{dressler2019data,dressler2020towards} & Deep Squats & 1 & \checkmark & & - & \checkmark\\
 \hline
  \cite{hu2014real} & Qigong+others & 4+6 & \checkmark & &  & \checkmark\\
 \hline
  \cite{capecci2016physical,capecci2018hidden} & Physiotherapy & 5 & \checkmark & &  & \checkmark\\
 \hline
  \cite{osgouei2018objective} & Shoulder Abduction & 1 & \checkmark & &  & \\
 \hline
 \cite{al2019quantifying} & General & 3 & & \checkmark  &  & \checkmark \\
 \hline
 \cite{cao2019novel} & Brunnstrom Scale & 9 & & \checkmark  &  &  \\
 \hline
 \cite{williams2019assessment} & Rehabilitation & 2 & \checkmark &  &  &  \\
 \hline
 \cite{yu2019dynamic} & Tai Chi & 1 & \checkmark &  &  & \checkmark \\
 \hline
 \cite{lei2020learning} & Olympic Sports & 9 & \checkmark & & &  \\
 \hline
\end{tabular}}
%\label{table:tblFeatures}
\\
\caption[]{Features of works that deal with assessing structured movements. The minus sign represents missing information.}
\label{tbl:features}
%\caption{}
\end{table}

\subsection{Public Datasets}
Many of the works used existing public datasets for evaluating their solutions, while others created their own datasets, for different assessment tasks. The used datasets have either been kept private or made public~\cite{paiement2014online,nguyen2018estimating,nguyen2018skeleton,chaaraoui2015abnormal}. Some of the works used both public and private datasets. Table~\ref{tbl:datasets} lists the public datasets used by existing works.

\begin{table}

\centering
\resizebox{0.9\linewidth}{!}{%
\begin{tabular}{ |c|c|c| } 
 \hline
 \textbf{Dataset} & \textbf{Movement Types} & \textbf{Used by} \\
 \hline
 SPHERE-staircase 2014,2015~\cite{paiement2014online} & Gait on stairs & \cite{paiement2014online,chaaraoui2015abnormal,devanne2016learning,baptista2018deformation,khokhlova2018kinematic}  \\
 \hline
 DGD: DAI gait dataset~\cite{chaaraoui2015abnormal} & Gait & \cite{chaaraoui2015abnormal,devanne2016learning,khokhlova2018kinematic}  \\
 \hline
 Walking gait dataset~\cite{nguyen2018walking} & Gait, under 9 different conditions  & \cite{jun2020feature,nguyen2018estimating,nguyen2018skeleton,khokhlova2018kinematic}  \\
 \hline
 UPCV Gait K2~\cite{kastaniotis2016pose} & Gait - normal walking  & \cite{khokhlova2018kinematic}  \\
 \hline
 Eyes. Mocap data~\cite{eyesmocapdata} & Gait captured by a Mocap system  & \cite{nguyen2016skeleton}  \\
 \hline
 HMRA~\cite{hmra} & Qigong and others  & \cite{hu2014real}  \\
 \hline
 UI-PRMD~\cite{vakanski2018data} & Physical therapy & \cite{williams2019assessment} \\
 \hline
 MIT Olympic Scoring Dataset~\cite{mitolympic} & Olympic scoring on RGB videos  & \cite{lei2020learning}  \\
 \hline
 UNLV Olympic Scoring Dataset~\cite{unlvoplymic} & Olympic scoring on RGB videos  & \cite{lei2020learning}  \\
 \hline
\end{tabular}}
%\label{table:tblFeatures}
\\
\caption[]{Public movement assessment datasets.}
\label{tbl:datasets}
%\caption{}
\end{table}

\subsection{Methods and Algorithms}

\subsubsection{Skeleton Detection.}
\begin{figure}[]
\centering
  \includegraphics[width=0.2\linewidth,keepaspectratio]{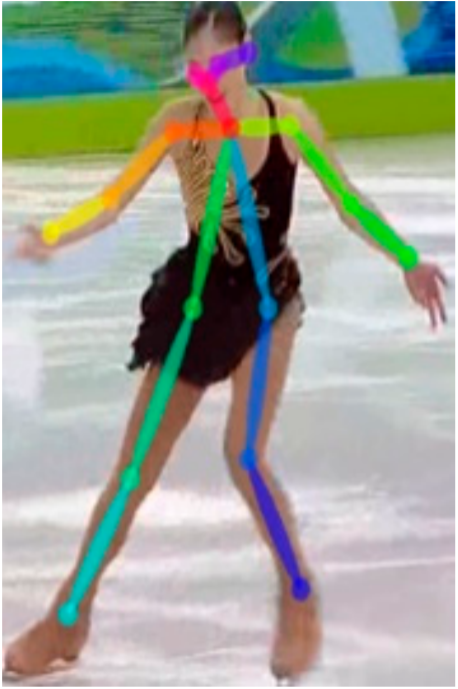}
  \caption[]{An OpenPose 2D skeleton~\cite{lei2020learning}.}
  \label{fig:openpose}
\end{figure}

The majority of the works use 3D cameras, such as Kinect1 or Kinect2, with the Microsoft Kinect SDK~\cite{shotton2011real} or OpenNI for detection of 3D skeletons. Sometimes, marker-based motion-capture (Mocap) systems are used~\cite{nguyen2016skeleton,al2019quantifying,williams2019assessment}. Lei \etal~\cite{lei2020learning} used 2D skeletons that were extracted from RGB videos, using OpenPose~\cite{cao2017realtime}, as visualized in Figure~\ref{fig:openpose}.

\subsubsection{Geometric Normalization.}
\begin{figure}[]
\centering
  \includegraphics[width=0.35\linewidth,keepaspectratio]{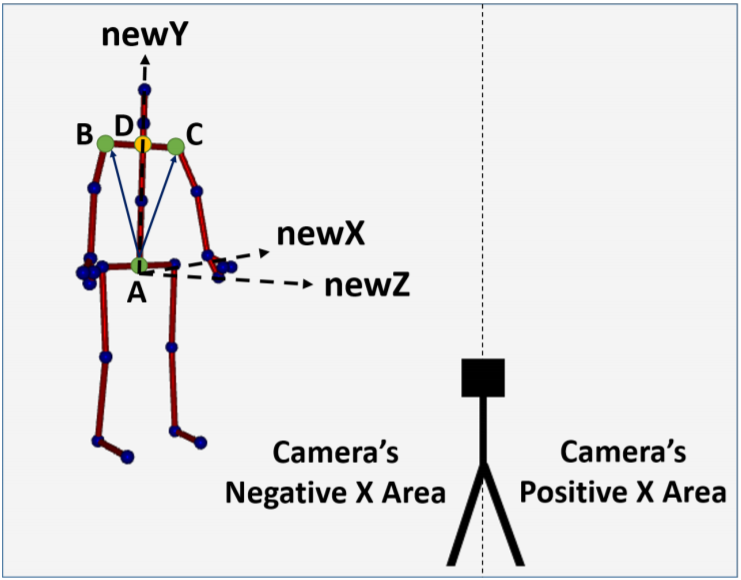}
  \caption[]{A geometric normalization step~\cite{hakim2019mal,hakim2019malf}.}
  \label{fig:geometric}
\end{figure}

%\label{lbl:geometric}
People perform movements in different distances and angles in respect to the 3D camera that captures their motion. Additionally, different people have different body dimensions, which have to be addressed by either pre-normalizing the skeleton dimensions and coordinates, as demonstrated in Figure~\ref{fig:geometric}, or extracting features that are inherently invariant to the camera location and body-dimensions, such as joint angles. This step therefore, may be considered an either independent or integral part of the feature-extraction process. Table~\ref{tbl:geometric} summarizes the geometric normalization methods used by existing works.

\begin{table}

\centering
\resizebox{0.98\linewidth}{!}{%
\begin{tabular}{ |c|l| } 
 \hline
 \textbf{Work} & \textbf{Implementation} \\
 \hline
 \cite{paiement2014online} & Translation, rotation and scaling due to varying heights of the subjects. \\
 \hline
 \cite{jun2020feature} &  Implementing the method from~\cite{paiement2014online}. \\
 \hline
 \cite{chaaraoui2015abnormal} & Translation, rotation by shoulder and hip joints, scaling.\\
 \hline
 \cite{nguyen2016skeleton} & Using features that are invariant to camera location and angle. \\
 \hline
 \cite{devanne2016learning} & - \\
 \hline
 \cite{baptista2018deformation} & Projection on the main direction of the motion variation. \\
 \hline
 \cite{nguyen2018estimating,nguyen2018skeleton} &  Scaling the coordinates to the range between 0 and 1. \\
 \hline
  \cite{khokhlova2018kinematic} & Using features that are invariant to camera location and angle. \\
 \hline
 \cite{parisi2016human} &  Translation. \\
 \hline
 \cite{su2013personal} & Geometric calibration as a system initialization step, before capturing skeleton videos. \\
 \hline
 \cite{eichler20183d,eichler2018non} & Using features that are invariant to camera location and angle. \\
 \hline
 \cite{hakim2019mal,hakim2019malf} & Projection on spine-shoulders plane, translation and equalizing skeleton edge lengths. \\
 \hline
 \cite{masalha2020predicting} & Using features that are invariant to camera location and angle. \\
 \hline
 \cite{dressler2019data,dressler2020towards} & Using features that are invariant to camera location and angle. \\
 \hline
 \cite{hu2014real} & Using features that are invariant to camera location and angle. \\
 \hline
 \cite{capecci2016physical,capecci2018hidden} & Using features that are invariant to camera location and angle. \\
 \hline
 \cite{osgouei2018objective} & Using features that are invariant to camera location and angle. \\
 \hline
 \cite{al2019quantifying} & Using features that are invariant to camera location and angle. \\
 \hline
 \cite{cao2019novel} & - \\
 \hline
 \cite{williams2019assessment} & Using features that are invariant to camera location and angle. \\
 \hline
 \cite{yu2019dynamic} & Projection on arm-leg-based coordinate system. \\
 \hline
 \cite{lei2020learning} & Scaling of the 2D human body. \\
 \hline

\end{tabular}}
\caption[]{Geometric normalization methods.}
\label{tbl:geometric}
\end{table}

\subsubsection{Temporal Alignment.}
\begin{figure}[]
\centering
  \includegraphics[width=0.45\linewidth,keepaspectratio]{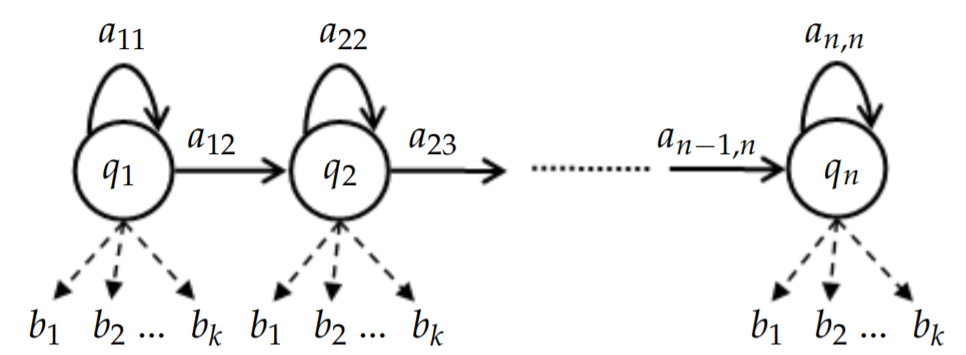}
  \caption[]{A Hidden Markov Model (HMM), which defines states, observations and probabilities of state transitions and observations~\cite{nguyen2016skeleton}.}
  \label{fig:hmm}
\end{figure}

\begin{figure}[]
\centering
  \includegraphics[width=0.65\linewidth,keepaspectratio]{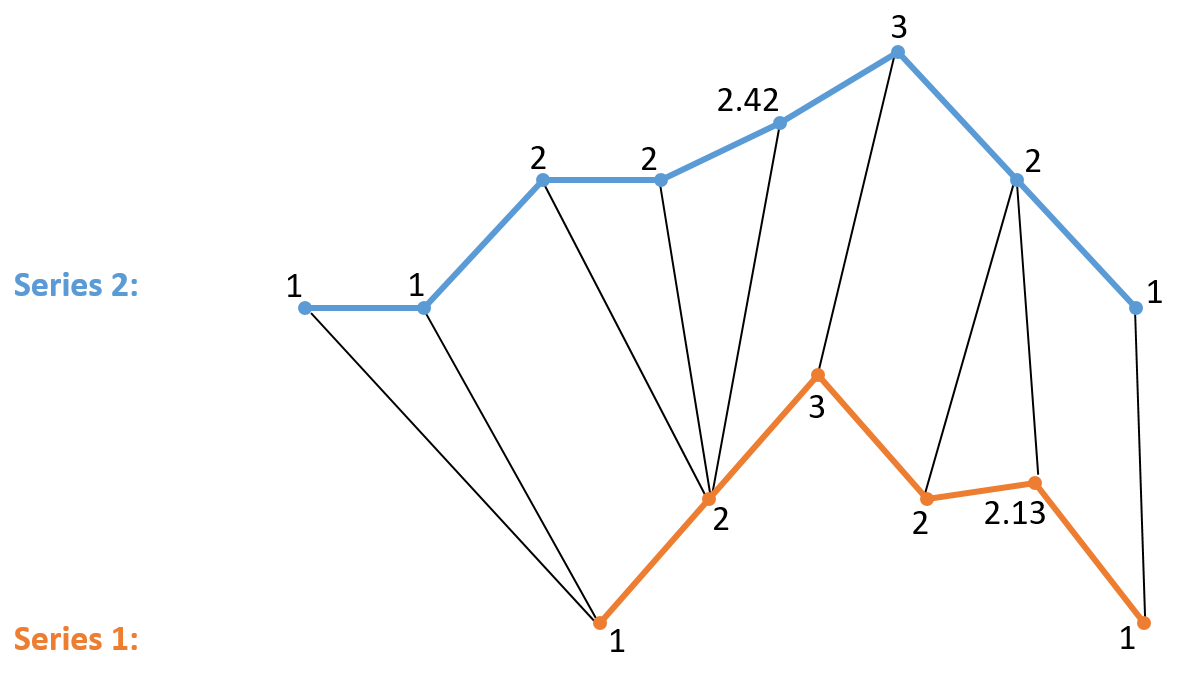}
  \caption[]{Dynamic Time Warping (DTW) for alignment of two series of scalars, by matching pairs of indices~\cite{simpledtw}.}
  \label{fig:dtw}
\end{figure}

%\label{lbl:temporalAlignment}
In order to produce reliable assessment outputs, a tested movement, which is a temporal sequence of data, usually has to be well-aligned in time with movements it will be compared to. For that purpose, most works either use models that inherently deal with sequences, such as HMMs and RNNs, as illustrated in Figure~\ref{fig:hmm}, or use temporal alignment algorithms, such as the DTW algorithm or its variants, as illustrated in Figure~\ref{fig:dtw}. 

\begin{figure}[]
\centering
  \includegraphics[width=0.8\linewidth,keepaspectratio]{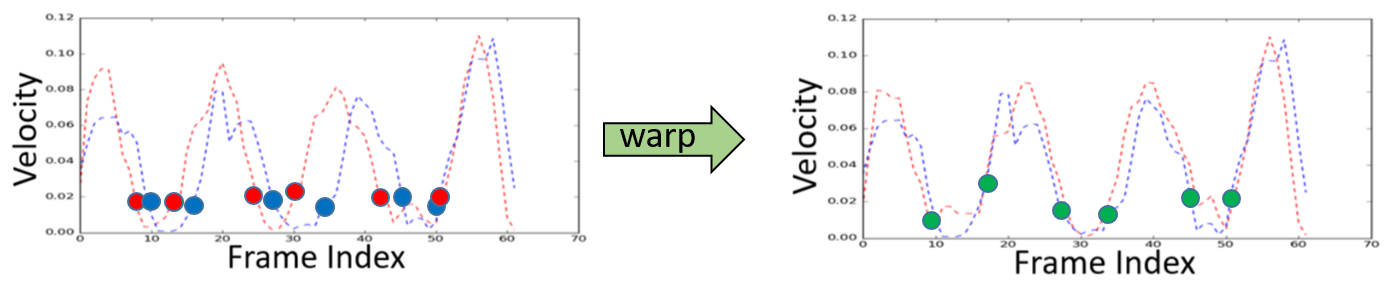}
  \caption[]{A continuous warping by scaling between detected pairs of temporal points-of-interest~\cite{hakim2019mal,hakim2019malf}.}.
  \label{fig:warp}
\end{figure}

Hakim and Shimshoni~\cite{hakim2019mal,hakim2019malf} introduced a novel warping algorithm, which was based on the detection of temporal points-of-interest (PoIs) and on linearly warping the sequences between them, as illustrated in Figure~\ref{fig:warp}. Dressler \etal~\cite{dressler2019data,dressler2020towards} introduced a novel DTW variation with skips, similarly to Hu \etal~\cite{hu2014real}. Other novel approaches were introduced by Devanne \etal~\cite{devanne2016learning}, by Baptista \etal~\cite{baptista2018deformation} and by Yu and Xiong~\cite{yu2019dynamic}.
Another less mentioned algorithm is the Correlation Optimized Warping
(COW) algorithm~\cite{tomasi2004correlation}.
Palma \etal~\cite{palma2016hmm} and Hagelb\"{a}ck \etal~\cite{hagelback2019variants} elaborated more on the topic of temporal alignment algorithms in the context of movement assessment. Table~\ref{tbl:temporal} summarizes the alignment methods used by existing works.

\begin{table}

\centering
\resizebox{0.8\linewidth}{!}{%
\begin{tabular}{ |c|l| } 
 \hline
 \textbf{Work} & \textbf{Implementation} \\
 \hline
 \cite{paiement2014online} & Inherently solved by the choice to use an HMM statistical model.  \\
 \hline
 \cite{jun2020feature} & Inherently solved by the choice to use an RNN Autoencoder. \\
 \hline
 \cite{chaaraoui2015abnormal} & Discrete warping using the Dynamic Time Warping (DTW) algorithm. \\
 \hline
 \cite{nguyen2016skeleton} & Inherently solved by the choice to use an HMM statistical model. \\
 \hline
 \cite{devanne2016learning} & Riemannian shape analysis of legs shape evolution within a sliding window. \\
 \hline
 \cite{baptista2018deformation} & Key-point detection with deformation-based curve alignment~\cite{demisse2017deformation}. \\
 \hline
 \cite{nguyen2018estimating,nguyen2018skeleton} &  Inherently solved by the choice to use a recurrent neural network.\\
 \hline
  \cite{khokhlova2018kinematic} & - \\
 \hline
 \cite{parisi2016human} & Inherently solved by the choice to use a recurrent neural network. \\
 \hline
 \cite{su2013personal} & Discrete warping using the Dynamic Time Warping (DTW) algorithm. \\
 \hline
 \cite{eichler20183d,eichler2018non} & - \\
 \hline
 \cite{hakim2019mal,hakim2019malf} & Detecting mutual temporal PoIs and continuously warping between them. \\
 \hline
 \cite{masalha2020predicting} & - \\
 \hline
 \cite{dressler2019data,dressler2020towards} & A novel DTW variant, with skips. \\
 \hline
 \cite{hu2014real} & A novel DTW variant with tolerance to editing. \\
 \hline
 \cite{capecci2016physical,capecci2018hidden} & DTW and Hidden Semi-Markov Models (HSMM). \\
 \hline
 \cite{osgouei2018objective} & DTW and HMM. \\
 \hline
 \cite{al2019quantifying} & - \\
 \hline
 \cite{cao2019novel} &  Inherently solved by the choice to use a recurrent neural network. \\
 \hline
 \cite{williams2019assessment} & - \\
 \hline
 \cite{yu2019dynamic} & A novel DTW variant that minimizes angles between pairs of vectors. \\
 \hline
 \cite{lei2020learning} & - \\
 \hline

\end{tabular}}
\caption[]{Temporal alignment methods.}
\label{tbl:temporal}
\end{table}

\subsubsection{Feature Extraction.}
\begin{figure}[]
\centering
  \includegraphics[width=0.6\linewidth,keepaspectratio]{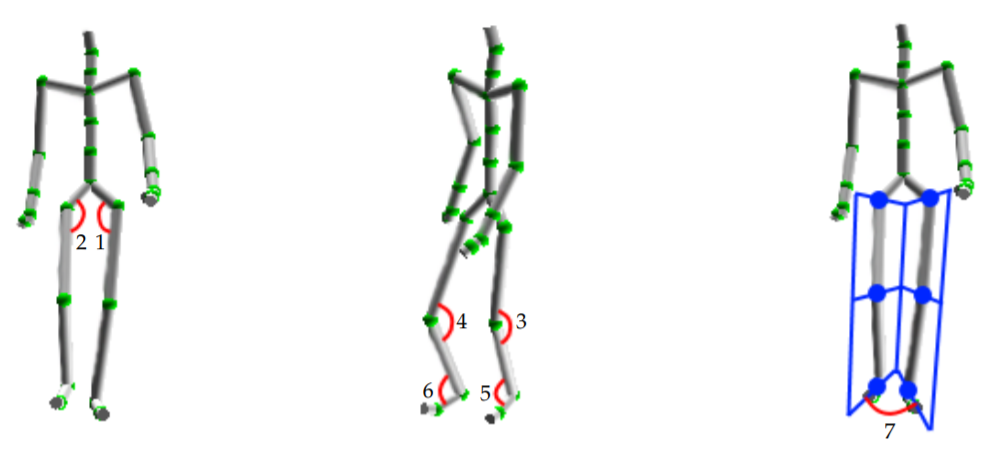}
  \caption[]{Angles as extracted skeleton features, which are invariant to the camera location and to body dimension differences~\cite{nguyen2016skeleton}.}.
  \label{fig:feature}
\end{figure}

The assessment of different types of movements requires paying attention to different details, which may include joint angles, pairwise joint distances, joint positions, joint velocities and event timings. Many of the feature extraction methods are invariant to the subject's skeleton scale and to the camera location and angle, as illustrated in Figure~\ref{fig:feature}, while others are usually preceded by a geometric normalization step. In the recent years, some works used deep features, which were automatically learned and were obscure, rather than using explainable handcrafted features.

It is notable that while some works were designated for specific domains of movements and exploited their prior knowledge to choose their features, other works were designated to be more versatile and adaptive to many movement domains, and therefore used general features. Table~\ref{tbl:feature} summarizes the feature extraction methods used by existing works.

\begin{table}
\centering
\resizebox{0.98\linewidth}{!}{%
\begin{tabular}{ |c|l| } 
 \hline
 \textbf{Work} & \textbf{Implementation} \\
 \hline
 \cite{paiement2014online} & Applying Diffusion Maps~\cite{coifman2006diffusion} on the normalized 3D skeleton joint positions. \\
 \hline
 \cite{jun2020feature} & Deep features learned by training RNN Autoencoders. \\
 \hline
 \cite{chaaraoui2015abnormal} & Joint Motion History (JMH): spatio-temporal joint 3D positions. \\
 \hline
 \cite{nguyen2016skeleton} & Lower-body joint angles and the angle between the two feet. \\
 \hline
 \cite{devanne2016learning} & Square-root-velocity function (SRVF)~\cite{joshi2007novel} on temporal sequences of joint positions. \\
 \hline
 \cite{baptista2018deformation} & Distances between the projections of the two knees on the movement direction. \\
 \hline
 \cite{nguyen2018estimating,nguyen2018skeleton} &  Deep features learned by Autoencoders. \\
 \hline
  \cite{khokhlova2018kinematic} & Covariance matrices of hip and knee flexion angles. \\
 \hline
 \cite{parisi2016human} & 13 joint 3D positions and velocities. \\
 \hline
 \cite{su2013personal} & Joint 3D positions and velocities. \\
 \hline
 \cite{eichler20183d,eichler2018non} & Joint angles, distances and heights from the ground. \\
 \hline
 \cite{hakim2019mal,hakim2019malf} & Joint 3D positions and velocities, distances and edge angles. Sequence timings. \\
 \hline
 \cite{masalha2020predicting} & Relative joint positions, joint distances, angles and height of joints from the ground. \\
 \hline
 \cite{dressler2019data,dressler2020towards} &  Joint positions and NASM features (a list of selected skeleton angles). \\
 \hline
 \cite{hu2014real} & Torso direction and joint relative position represented by elevation and azimuth. \\
 \hline
 \cite{capecci2016physical,capecci2018hidden} & Selected features varying between movement types. \\
 \hline
 \cite{osgouei2018objective} & Shoulder and arm angles. \\
 \hline
 \cite{al2019quantifying} & Autoencoder embeddings of manually-extracted attributes. \\
 \hline
 \cite{cao2019novel} & Raw skeleton 3D data. \\
 \hline
 \cite{williams2019assessment} & GMM encoding of Autoencoder dimensionality-reduced joint angle data. \\
 \hline
 \cite{yu2019dynamic} & Angles of selected joints. \\
 \hline
 \cite{lei2020learning} &  Self-similarity descriptors of joint trajectories and a joint displacement sequence.\\
 \hline

\end{tabular}}
\caption[]{Feature extraction methods.}
\label{tbl:feature}
\end{table}

\subsubsection{Score Prediction.}
The prediction of an assessment score refers to one or more of the following cases:
\begin{enumerate}
    \item Classifying a performance into a class from a predefined set of discrete quality classes.
    \item Performing a regression that will map a performance into a number on a predefined continuous scale.
    \item Producing scores that reflect the similarity between given model and performance, unbound to ground-truth or predefined scales.
\end{enumerate}

\noindent The two first types of scoring capabilities are mainly essential for formal assessments, such as medical assessments or Olympic performance judgements. The third type of scoring is mainly useful for comparing subject performances, which can be either a certain subject whose progress is monitored over time, or different subjects who compete.

Table~\ref{tbl:score} lists the algorithms used to produce quality scores. It is notable that score prediction was not implemented in many works, as it was irrelevant for them, since they only addressed normal/abnormal binary classifications.

\begin{table}
\centering
\resizebox{0.98\linewidth}{!}{%
\begin{tabular}{ |c|l| } 
 \hline
 \textbf{Work} & \textbf{Implementation} \\
 \hline
 \cite{paiement2014online} & Pose and dynamics log likelihoods. \\
 \hline
 \cite{jun2020feature} & - \\
 \hline
 \cite{chaaraoui2015abnormal} & - \\
 \hline
 \cite{nguyen2016skeleton} & - \\
 \hline
 \cite{devanne2016learning} & Mean log-probability over the segments of the test sequence. \\
 \hline
 \cite{baptista2018deformation} & Distance between time-aligned feature sequences with reflection of time-variations. \\
 \hline
 \cite{nguyen2018estimating,nguyen2018skeleton} & - \\
 \hline
  \cite{khokhlova2018kinematic} & - \\
 \hline
 \cite{parisi2016human} & Difference between actual and RNN-predicted next frames.  \\
 \hline
 \cite{su2013personal} & Handcrafted classification using Fuzzy Logic~\cite{zadeh1965fuzzy}. \\
 \hline
 \cite{eichler20183d,eichler2018non} & SVM, Decision Tree and Random Forest quality classification using handcrafted features. \\
 \hline
 \cite{hakim2019mal,hakim2019malf} & Thresholded weighted sum of normalized, time-filtered active/inactive joint and timing scores. \\
 \hline
 \cite{masalha2020predicting} & SVM and Random Forest quality classification using handcrafted features. \\
 \hline
 \cite{dressler2019data,dressler2020towards} & Weighted sum of selected feature differences.  \\
 \hline
 \cite{hu2014real} & Average of frame cross-correlations. \\
 \hline
 \cite{capecci2016physical,capecci2018hidden} & Normalized log-likelihoods or DTW distances. \\
 \hline
 \cite{osgouei2018objective} & Difference from proper performance divided by difference between worst and proper performances. \\
 \hline
 \cite{al2019quantifying} & Classification using One-Class SVM. \\
 \hline
 \cite{cao2019novel} & Classification using a hybrid LSTM-CNN model. \\
 \hline
 \cite{williams2019assessment} & Normalized log-likelihoods. \\
 \hline
 \cite{yu2019dynamic} & DTW similarity. \\
 \hline
 \cite{lei2020learning} & Regression based on high-level features combined with joint trajectories and displacements. \\
 \hline

\end{tabular}}
\caption[]{Score prediction methods.}
\label{tbl:score}
\end{table}

\subsubsection{Feedback Generation.}
\begin{figure}[]
\centering
  \includegraphics[width=0.75\linewidth,keepaspectratio]{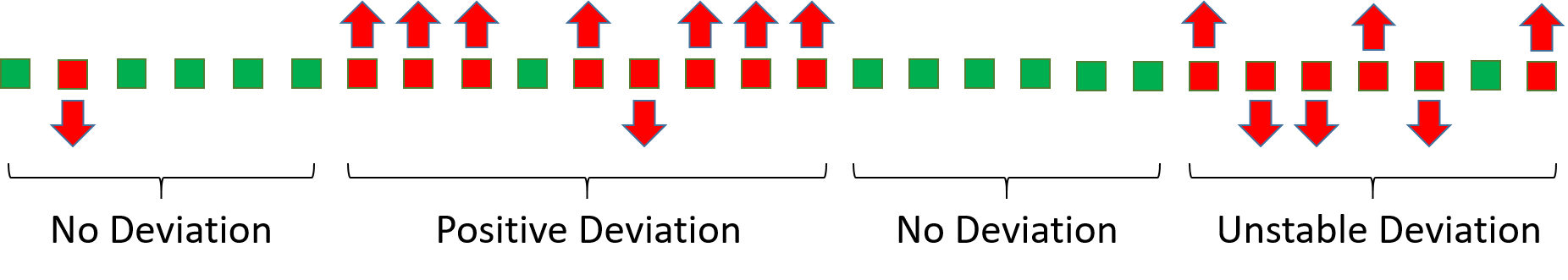}
  \caption[]{Temporal segmentation of parameter deviations for feedback generation~\cite{hakim2019mal,hakim2019malf}.}
  \label{fig:ParameterTimeSegmentation}
\end{figure}

There are two main types of feedback that can be generated: bound feedback and unbound feedback. Feedback is bound when it can only consist of predefined mistakes or abnormalities that can be detected. Feedback is unbound when it is a generated natural language text that can describe any type of mistake, deviation or abnormality. The generation of unbound feedback usually requires the usage of describable low-level features, so that when a performance is not proper, it will be possible to indicate the most significant features that reduced the score and translate them to natural language, such that the user can use the feedback to learn how to improve their next performance. Such feedback may include temporal sequences that deviate similarly, as visualized in Figure~\ref{fig:ParameterTimeSegmentation}.

Table~\ref{tbl:feedback} summarizes the types of feedback and generation methods used by the works. It is notable that: 1) Most of the works do not generate feedback. 2) There are no works that produce feedback while not predicting quality scores, for the same reason of only focusing on binary detection of abnormalities.

\begin{table}
\centering
\resizebox{0.9\linewidth}{!}{%
\begin{tabular}{ |c|l| } 
 \hline
 \textbf{Work} & \textbf{Implementation} \\
 \hline
 \cite{paiement2014online} & - \\
 \hline
 \cite{jun2020feature} & - \\
 \hline
 \cite{chaaraoui2015abnormal} & - \\
 \hline
 \cite{nguyen2016skeleton} & - \\
 \hline
 \cite{devanne2016learning} & Visualizing the deviations of the body parts. \\
 \hline
 \cite{baptista2018deformation} & - \\
 \hline
 \cite{nguyen2018estimating,nguyen2018skeleton} & - \\
 \hline
  \cite{khokhlova2018kinematic} & - \\
 \hline
 \cite{parisi2016human} & Sequences of parameter deviations and detection of predefined typical mistakes.  \\
 \hline
 \cite{su2013personal} & Three quality classifications indicating trajectory similarity and right speed. \\
 \hline
 \cite{eichler20183d,eichler2018non} & - \\
 \hline
 \cite{hakim2019mal,hakim2019malf} & Translation of worst class-segmented parameter temporal deviations into text. \\
 \hline
 \cite{masalha2020predicting} & - \\
 \hline
 \cite{dressler2019data,dressler2020towards} & Indication of weak links according to angle differences.  \\
 \hline
 \cite{hu2014real} & - \\
 \hline
 \cite{capecci2016physical,capecci2018hidden} & - \\
 \hline
 \cite{osgouei2018objective} & - \\
 \hline
 \cite{al2019quantifying} & - \\
 \hline
 \cite{cao2019novel} &  - \\
 \hline
 \cite{williams2019assessment} & - \\
 \hline
 \cite{yu2019dynamic} & - \\
 \hline
 \cite{lei2020learning} & - \\
 \hline

\end{tabular}}
\caption[]{Feedback generation methods.}
\label{tbl:feedback}
\end{table}

\section{Discussion}

From the reviewed works, we can learn that a typical movement assessment solution may deal with detection of abnormal events or predicting quality scores, using classification, regression or computation of a normalized similarity measure. The task of detecting abnormal events is usually associated with repetitive movements, while the task of predicting scores is usually associated with structured movements.

We can learn that while public skeleton datasets exist and are used by some of the works, most of the works use private datasets that were acquired for the sake of a specific work. It is notable that while many novelties are proposed in the different works, the absence of common datasets and evaluation metrics leads to different works dealing with different problems, evaluating themselves on different datasets of different movement domains, using different metrics.

It is notable that temporal alignment is a key-problem in movement assessment. From the reviewed works, we can learn that around a half of the works base their temporal alignment solutions on models that are designated for sequential inputs, such as Hidden Markov Models and recurrent neural networks, while others use either the Dynamic Time Warping algorithm, sometimes with novel improvements, or other novel warping and alignment approaches.

We can learn that while a few works use features that were automatically learned by neural networks, most of the works make use of handcrafted skeleton features. In many of those works, the use features are invariant to the camera location and angle and to the body-dimensions of the performing subjects. Other works that make use of handcrafted features usually have to apply a geometric normalization step before continuing to the next steps. It is worth to mention that while some of the works were designed to deal with a specific type of movement, other works were designed to be adaptive and deal with many types of movements, a choice that is usually clearly reflected in the feature extraction step.

We can learn that a quality score is produced by most of the works. While works that deal with medical assessments mainly focus on classification into predefined discrete scoring scales, other works predict scores on continuous scales. Such scores are rarely learned as a regression problem and are usually based on a normalized similarity measure. Finally, we can learn that only a few works deal with producing feedback, which can be bound or unbound.

In the future, the formation of a large, general public dataset and a common evaluation metric may help define the state-of-the-art and boost the research on the topic of movement assessment. In addition, the improvement of mobile-device cameras, as well as computer vision algorithms that detect skeletons in RGB-D or even RGB videos, may raise the interest in researching this topic.

\section{Conclusions}
We have provided a review of the existing works in the domain of movement assessment from skeletal data, which gives a high level picture of the problems addressed and approaches implemented by existing solutions.

We divided the types of assessment tasks into two main categories, which were detection of abnormalities in long, repetitive movements and scoring structured movements, sometimes while generating textual feedback. The objectives and challenges of the assessment task were discussed and the ways they were addressed by each of the works were listed, including skeleton joint detection, geometric normalization, temporal alignment, feature extraction, score prediction and feedback generation. The existing public datasets and evaluated movement domains were listed. Finally, a high level discussion was provided. We hope that this review will provide a good starting point for new researchers.

\bibliographystyle{splncs}
\bibliography{egbib}

%this would normally be the end of your paper, but you may also have an appendix
%within the given limit of number of pages
\end{document}